\documentclass[letterpaper, 10 pt, conference]{ieeeconf}  

\IEEEoverridecommandlockouts                              
                                                          
\usepackage{lipsum} 
\usepackage{amsmath,amsfonts}
\usepackage{algorithmic}
\usepackage{algorithm}
\usepackage{array}
\usepackage[caption=false,font=normalsize,labelfont=sf,textfont=sf]{subfig}
\usepackage{textcomp}
\usepackage{stfloats}
\usepackage{url}
\usepackage{verbatim}
\usepackage{graphicx}
\usepackage{cite}
\usepackage{comment}
\usepackage{multirow}
\usepackage{xcolor}
\usepackage{amssymb}
\usepackage{booktabs}
\usepackage[hidelinks]{hyperref}\usepackage{etoolbox,xspace}
\usepackage{cuted}
\usepackage{capt-of}
\let\labelindent\relax
\usepackage{enumitem}

\def\shownotes{1}

\ifnum\shownotes=1
\newcommand\zhiming[1]{\textcolor{cyan}{Zhiming: #1}}

\newcommand\syn[1]{\textcolor{green}{Syn: #1}}
\newcommand\todo[1]{\textcolor{red}{#1}}
\newcommand\andreas[1]{\textcolor{orange}{Andreas: #1}}
\newcommand\daniel[1]{\textcolor{blue}{Daniel: #1}}
\else 
\newcommand\andreas[1]{}
\newcommand\syn[1]{}
\newcommand\zhiming[1]{}

\newcommand\todo[1]{}
\newcommand\daniel[1]{}
\fi

\author{Zhiming Hu, Syn Schmitt, Daniel Häufle, and Andreas Bulling
\thanks{Zhiming Hu, Syn Schmitt, and Andreas Bulling are with the University of Stuttgart, Germany. E-mail: \{zhiming.hu@vis.uni-stuttgart.de, schmitt@simtech.uni-stuttgart.de, andreas.bulling@vis.uni-stuttgart.de\}. Daniel Haeufle is with Heidelberg University and University of Tuebingen, Germany. E-mail: daniel.haeufle@ziti.uni-heidelberg.de. 
Syn Schmitt, Daniel Haeufle, and Andreas Bulling are with the Center for Bionic Intelligence Tuebingen Stuttgart (BITS), Germany. %
Zhiming Hu is the corresponding author.
This work was funded by the Deutsche Forschungsgemeinschaft (DFG, German Research Foundation) under Germany's Excellence Strategy -- EXC 2075 -- 390740016.
}}

\begin{document}
\newcommand{\methodName}{GazeMotion\xspace}
\title{\LARGE \bf \methodName: Gaze-guided Human Motion Forecasting}

\maketitle
\thispagestyle{empty}
\pagestyle{empty}



\begin{abstract}
We present \textit{\methodName}~-- a novel method for human motion forecasting that combines information on past human poses with human eye gaze.
Inspired by evidence from behavioural sciences showing that human eye and body movements are closely coordinated,
\methodName first predicts future eye gaze from past gaze, then fuses predicted future gaze and past poses into a gaze-pose graph, and finally uses a residual graph convolutional network to forecast body motion.
We extensively evaluate our method on the MoGaze, ADT, and GIMO benchmark datasets and show that it outperforms state-of-the-art methods by up to 7.4\% improvement in mean per joint position error.
Using head direction as a proxy to gaze, our method still achieves an average improvement of 5.5\%.
We finally report an online user study showing that our method also outperforms prior methods in terms of perceived realism.
These results show the significant information content available in eye gaze for human motion forecasting as well as the effectiveness of our method in exploiting this information.
\end{abstract}

\section{Introduction}

Understanding and forecasting human motion -- coarse activities such as walking, or fine-grained movements such as reaching or grasping -- is a long-standing research challenge in mobile robotics and human-robot interaction~\cite{belardinelli2022intention, shi2021gazeemd}.
Given the inherent sequential nature of human motion, much previous work on motion forecasting has focused on using recurrent neural networks, showing significant performance improvements~\cite{le2021hierarchical, martinez2017human}.
Other methods for human motion forecasting include Transformer-based architectures~\cite{mao2021multi}, graph convolutional networks (GCNs)~\cite{ma2022progressively}, or
multi-layer perceptrons (MLPs)
\cite{guo2023back}.
Common to all of these methods is that they formulate motion forecasting as a sequence-to-sequence task in which future motion is predicted \textit{solely} from past human movements or poses~\cite{ma2022progressively, mao2021multi, martinez2017human, guo2023back}.

In a parallel line of work, studies in the cognitive and behavioural sciences have revealed the strong correlation between human eye gaze and body movements during daily activities~\cite{freedman2008coordination, goossens1997human}.
For example, when navigating, human visual attention is strongly correlated with the movement direction~\cite{sun2018towards} and it tends to precede corresponding hand and arm movements when reaching for an object~\cite{emery2021openneeds}.
Despite the close coordination between human eye gaze and body movements, this information has only recently started to be explored for human motion forecasting. A method proposed in~\cite{zheng2022gimo} requires rich information about the full 3D environment and objects therein.

We present \textit{\methodName} -- the first learning-based method for human motion forecasting that combines information on past human poses with eye gaze without requiring such information.
Our method consists of three main components: a convolutional neural network to predict future eye gaze from historical gaze, a pose-gaze graph that fuses pose and gaze data, and a novel residual graph convolutional network consisting of three spatial-temporal modules that forecasts future body poses from the pose-gaze graph.
We evaluate our method for motion forecasting at different future time horizons of up to $1$ second (future 30 frames) on the MoGaze~\cite{kratzer2020mogaze}, ADT~\cite{pan2023aria}, and GIMO~\cite{zheng2022gimo} benchmark datasets.
We show that our method outperforms several state-of-the-art methods by a large margin: We achieve 7.4\% improvement on MoGaze, 6.4\% on ADT, and 6.1\% on GIMO in terms of mean per joint position error (MPJPE).
We further report an online user study
that shows that our method outperforms other methods in terms of precision and perceived realism of the predicted human motion.
Considering that eye gaze information is not always available in real applications, we further use head direction as a proxy to gaze and show that our method still achieves significantly better performances than prior methods.
The full source code and trained models are available at zhiminghu.net/hu24\_gazemotion.

\vspace{1em}
\noindent
The specific contributions of our work are three-fold:
\begin{itemize}

\item We propose a novel learning-based method that predicts future eye gaze from past gaze, fuses the predicted future gaze and past poses into a gaze-pose graph, and forecasts future poses through a novel residual graph convolutional network.

\item We report extensive experiments on three public datasets for motion forecasting at different future time horizons and demonstrate significant performance improvements over several state-of-the-art methods.

\item We conduct an online user study and validate that our method outperforms prior methods in both precision and realism.

\end{itemize}

\section{Related Work}

\subsection{Coordination of Eye and Body Movements}
The coordination of human eye and body movements has been extensively studied in multiple fields, including cognitive science and human-centred computing.
Hu et al.~\cite{hu2019sgaze, hu2021fixationnet, hu24pose2gaze} have demonstrated that head movements are strongly correlated with eye movements in many daily activities, such as free viewing or searching for an object.
Emery et al.~\cite{emery2021openneeds} investigated the coordination of eye, hand, and head movements in a virtual environment. 
In this work, we demonstrate that findings on the strong link between human motion and eye gaze can be successfully transferred to the task of motion forecasting and lead to significant performance improvements.

\subsection{Eye Gaze Prediction}

Human eye gaze is significant for many important applications including human-robot interaction~\cite{belardinelli2022intention, shi2021gazeemd} and human action anticipation~\cite{duarte2018action,hu2022ehtask}.
In this background, eye gaze prediction has become a popular research topic in the areas of robotics and human-centred computing.
Kim et al. focused on robot manipulation scenarios and proposed a Transformer-based method to predict eye gaze based on sequential visual input~\cite{kim2022memory}.
Hu et al. concentrated on a visual search task and used the scene content and task-related information to forecast future eye gaze~\cite{hu2021fixationnet}.
Existing methods usually rely on additional information, e.g. image content and task-related variables, making it difficult to apply their methods to other situations.
In contrast, our method only employs historical eye gaze to forecast future gaze.
In addition, we are the first to combine gaze prediction with motion forecasting, i.e., we first forecast future eye gaze and then use the predicted future gaze to forecast future motion.



\section{Method}

\begin{figure*}
\centering
    \includegraphics[width=\textwidth]{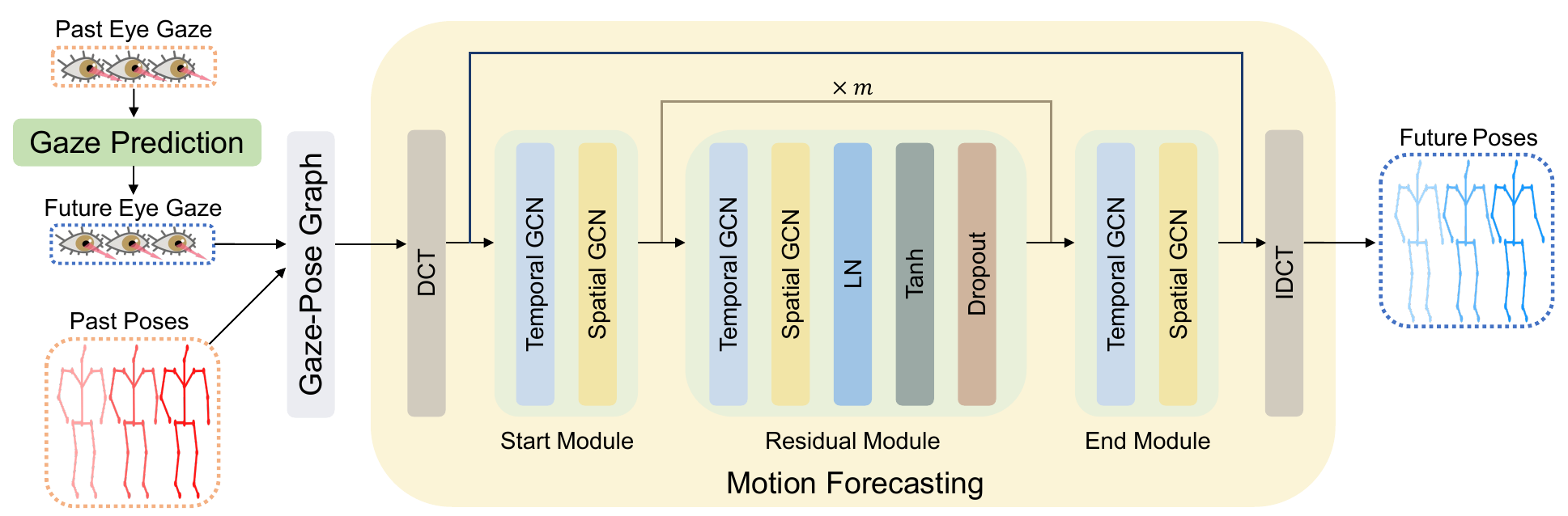}
    \caption{Our method 
    first forecasts future eye gaze from past gaze using a 1D convolutional neural network, then fuses the predicted gaze and past body poses into a gaze-pose graph, and finally applies a graph convolutional network consisting of a start, a residual, and an end module to forecast body motion.}
    \label{fig:method}
\end{figure*}


We define gaze-guided human motion forecasting as the task of predicting future body motion, for a certain time horizon, jointly from past body poses and eye gaze information. Human pose $p$ is represented by the 3D positions of all human joints $p\in R^{3\times n}$, where $n$ is the number of joints.
Eye gaze direction is indicated using a unit vector $g\in R^3$.
Given a sequence of past human poses $P_{1:t} = \{p_1, p_2, ..., p_t\}$ and human eye gaze $G_{1:t} = \{g_1, g_2, ..., g_t\}$, the task is to forecast human poses in the future $P_{t+1:T} = \{p_{t+1}, p_{t+2}, ..., p_{T}\}$.
Our method consists of three main components: a gaze prediction network that forecasts future eye gaze from past gaze, a gaze-pose fusion procedure that fuses predicted future gaze and past poses into a gaze-pose graph, and a motion forecasting network that forecasts future motions from the fused gaze and motion data (see Figure \ref{fig:method} for an overview of our method).

\subsection{Gaze Prediction}

Guided by the intuition that information on future eye gaze is more useful for motion forecasting than past eye gaze, we first forecast future gaze $G_{t+1:2t}$ from past gaze $G_{1:t}$ and then use the predicted future gaze to forecast future body motion.
In light of the good performance of 1D CNN for processing eye gaze data~\cite{hu2020dgaze,hu2021fixationnet}, we employed four 1D CNN layers to forecast eye gaze.
More specifically, we used three 1D CNN layers, each with $32$ channels and a kernel size of three, to extract features from the historical eye gaze data.
Each CNN layer was followed by a layer normalisation (LN) and a Tanh activation function.
After the three CNN layers, we used a 1D CNN layer with three channels and a kernel size of three, and a Tanh activation function to predict future eye gaze from the gaze features.
The predicted gaze directions were finally normalised to unit vectors: $\hat{G}_{t+1:2t} = \{\hat{g}_{t+1}, \hat{g}_{t+2}, ..., \hat{g}_{2t}\} \in R^{3\times t}$.

\subsection{Gaze-Pose Fusion} \label{sec:fusion}

We first padded the past poses and predicted gaze up to $T$ by repeating the last pose and gaze for $T-t$ times following prior works~\cite{ma2022progressively, mao2019learning}.
Each pose was represented using the 3D coordinates of all the human joints: $P_{1:T} = \{J^{1}_{1:T}, J^{2}_{1:T}, ..., J^{n}_{1:T}\}\in R^{3\times n \times T}$.
We further cloned the predicted eye gaze data $\hat{G}$ for $n-1$ times to make the gaze data have the same size as the motion data.
We then concatenated the motion data and gaze data along the spatial dimension and obtained $X\in R^{3\times2n\times T}$.
We modelled the fused motion and gaze data $X$ as fully-connected spatial and temporal graphs, jointly called the \textit{gaze-pose graph}, to learn the relationship between human eye gaze and human motion.
The spatial graph represents each joint $J^k$ and each gaze $\hat{G}$ as a separate node and thus contains $2n$ nodes: $n$ joint nodes and $n$ gaze nodes.
The temporal graph consists of $T$ nodes, corresponding to the fused data at different times: $X_1, X_2, ..., X_T$.
The spatial and temporal graphs are both fully-connected with their adjacency matrices measuring the weights between each pair of nodes.

\subsection{Motion Forecasting}

We first applied discrete cosine transform (DCT) ~\cite{guo2023back,ma2022progressively} to encode the fused data $X$ in the temporal domain using
DCT matrix $M_{dct} \in R^{T\times T}$:
\begin{equation} 
\begin{aligned}
    X_{d} = XM_{dct}.
\end{aligned}
\end{equation}
We then proposed three novel GCN modules, i.e., a start module that mapped $X_{d}\in R^{3\times2n\times T}$ into feature space, a residual module that extracted the features, and an end module that mapped the features to the original data space.

\textbf{Start Module:} 
The start module first used a temporal GCN to extract the temporal features from the transformed data.
To this end, the temporal GCN learned the weighted adjacency matrix $A^T \in R^{T\times T}$ of the fully-connected temporal graph
and calculated temporal convolution using
\begin{equation} 
\begin{aligned}
    X^{1}_{d} = X_d A^T.
\end{aligned}
\end{equation}
$X^{1}_{d}\in R^{3\times2n\times T}$ was then permuted to $X^{2}_{d}\in R^{T\times2n\times 3}$.
A weight matrix $W^{start}\in R^{3\times16}$ was used to convert the input node features ($3$ dimensions) to latent features ($16$ dimensions):
\begin{equation} 
\begin{aligned}
    X^{3}_{d} = X^{2}_{d} W^{start}.
\end{aligned}
\end{equation}
After the weight matrix, a spatial GCN was employed to extract the spatial features.
It learned the weighted adjacency matrix $A^S \in R^{2n\times 2n}$ of the fully-connected spatial graph
and performed spatial convolution using 
\begin{equation} 
\begin{aligned}
    X^{4}_{d} = A^S X^{3}_{d}.
\end{aligned}
\end{equation}
$X^{4}_{d}\in R^{T\times2n\times 16}$ was further permuted to $X^{5}_{d}\in R^{16\times2n\times T}$.
Considering that repeating important features is beneficial for motion forecasting~\cite{ma2022progressively,mao2019learning}, we copied the output of the start module along the temporal dimension ($R^{16\times2n\times T}\rightarrow R^{16\times2n\times 2T}$) and used it as input to the residual module.

\textbf{Residual Module:} The residual module consisted of $m$ GCN blocks with each block containing a temporal GCN that learned the temporal adjacency matrix $A^T_i \in R^{2T\times 2T}$, a weight matrix $W^{res}_i\in R^{16\times16}$ that extracted the latent features, a spatial GCN that learned the spatial adjacency matrix $A^S_i \in R^{2n\times 2n}$, a layer normalisation, a Tanh activation function, and a dropout layer with dropout rate $0.3$ to prevent overfitting.
$m$ was set to $16$ and a residual connection was added for each GCN block.
We cut the output of the residual module in half in the temporal dimension ($R^{16\times2n\times 2T}\rightarrow R^{16\times2n\times T}$) and input it to the end module.

\textbf{End Module:} The end module consisted of a temporal GCN that learned the temporal adjacency matrix, a weight matrix $W^{end}\in R^{16\times3}$ that mapped the latent features to $3$ dimensions, and a spatial GCN that learned the spatial adjacency matrix.
We added a global residual connection to improve the network flow.
The output of the end module $Y_d \in R^{3\times2n\times T}$ was converted back to the original representation space using an inverse discrete cosine transform (IDCT) matrix $M_{idct} \in R^{T\times T}$:
\begin{equation} 
\begin{aligned}
    Y = Y_dM_{idct}.
\end{aligned}
\end{equation}
The predicted future poses $\hat{P}_{t+1:T} \in R^{3\times n\times {T-t}}$ were obtained from the joint nodes in $Y \in R^{3\times2n\times T}$.

\subsection{Loss Function}

We trained the gaze prediction and motion forecasting networks separately.
For the gaze prediction network, we used the angular error between the predicted eye gaze direction $\hat{G}$ and the ground truth gaze direction $G$ as the loss function:
\begin{equation} 
\begin{aligned}
    \ell = \arccos(\hat{G} \cdot G).
\end{aligned}
\end{equation}

For the motion forecasting network, we used a combination of motion loss $\ell_m$ and velocity loss $\ell_v$ as our loss function $\ell$:
\begin{equation} 
\begin{aligned}
    \ell = \ell_m + \ell_v.
\end{aligned}\label{eq:loss}
\end{equation}
$\ell_m$ measures the mean per joint position error between the predicted future poses and the ground truth~\cite{ma2022progressively}:
\begin{equation} 
\begin{aligned}
\ell_m =\frac{1}{n(T-t)}\sum_{j=t+1}^T\sum_{k=1}^n\lVert \hat{p}_{j,k} -p_{j,k}\rVert^2,
\end{aligned}\label{eq:mpjpe}
\end{equation}
where $\hat{p}_{j,k} \in R^3$ represents the 3D coordinates of the $k^{th}$ joint at the future time of $j$ while $p_{j,k} \in R^3$ is the corresponding ground truth.
$\ell_v$ measures the mean per joint velocity error between the predicted future poses and the ground truth:
\begin{equation} 
\begin{aligned}
\ell_v =\frac{1}{n(T-t-1)}\sum_{j=t+1}^{T-1}\sum_{k=1}^n\lVert \hat{v}_{j,k} -v_{j,k}\rVert^2,
\end{aligned}
\end{equation}
where $\hat{v}_{j,k} \in R^3$ represents the velocity of the $k^{th}$ joint at the future time of $j$ while $v_{j,k} \in R^3$ is the corresponding ground truth.
The velocity is computed using the time difference: $\hat{v}_{j,k} = \hat{p}_{j+1,k} - \hat{p}_{j,k}$ and $v_{j,k} = p_{j+1,k} - p_{j,k}$.
\section{Experiments and Results}

\subsection{Datasets}


\paragraph{MoGaze dataset~\cite{kratzer2020mogaze}} The MoGaze dataset
contains human motion and eye gaze data recorded at $120$ Hz from six people performing \textit{pick} and \textit{place} actions.
We down-sampled the human pose and gaze data to $30$ Hz for simplicity~\cite{ma2022progressively, guo2023back} and represented human poses using the 3D coordinates of $21$ human joints.
We used a leave-one-person-out cross-validation: We trained on five participants and tested on the remaining one, repeated the experiment six times with a different participant for testing, and calculated the average performance across all six iterations.

\paragraph{ADT dataset~\cite{pan2023aria}}
The ADT dataset
contains $35$ sequences of human pose and gaze data performing various indoor activities including \textit{room decoration}, \textit{meal preparation}, and \textit{work}.
Each human pose consists of the 3D coordinates of $21$ human joints recorded at $30$ Hz.
For experiments on ADT, we randomly selected $25$ sequences for training and $10$ sequences for testing.

\paragraph{GIMO dataset~\cite{zheng2022gimo}} 
The GIMO dataset is collected in various indoor scenes
and contains motion and gaze data of $11$ participants performing daily activities.
Each human pose consists of the 3D coordinates of $23$ human joints recorded at $30$ Hz.
We used the default train/test split
~\cite{zheng2022gimo}: motion and gaze data from $12$ scenes were used for training and data from $14$ scenes ($12$ known scenes from the training set and two new environments) for testing.

\subsection{Evaluation Settings}
\paragraph{Evaluation Metric} 
As is common in human motion forecasting~\cite{zheng2022gimo,guo2023back,ma2022progressively}, we used the mean per joint position error (see Equation \ref{eq:mpjpe}) in millimeters as our metric to evaluate the different motion forecasting methods.

\paragraph{Baselines}
We compared our method with the following state-of-the-art baseline methods for motion forecasting:
\begin{itemize}[noitemsep,leftmargin=*]
    
    \item \textit{Res-RNN}~\cite{martinez2017human}: \textit{Res-RNN} is a RNN-based method that applies a residual connection between the input pose and output pose to improve performance.

    \item \textit{siMLPe}~\cite{guo2023back}: \textit{siMLPe} is a light-weight MLP-based method that applies discrete cosine transform and residual connections to improve performance.
    
    \item \textit{HisRep}~\cite{mao2021multi}: \textit{HisRep} is a Transformer-based method that extracts motion attention to capture the similarity between the current motion context and the historical motion sub-sequences.
    
    \item \textit{PGBIG}~\cite{ma2022progressively}: \textit{PGBIG} is a GCN-based method that employs a multi-stage framework to forecast human motions where each stage predicts an initial guess for the next stage.
\end{itemize}

\paragraph{Time Horizons of Input and Output Sequences}
For experiments on the MoGaze, ADT, and GIMO datasets (30 Hz), we used $10$ frames of data as input to forecast human poses in the future $30$ frames (i.e., up to one second into the future), following the common evaluation settings~\cite{ma2022progressively, mao2021multi}.

\paragraph{Implementation Details}
We trained the baseline methods from scratch using their default parameters.
To train our gaze prediction network, we used the Adam optimiser with an initial learning rate of $0.01$ that we then decayed by $0.9$ every epoch.
We used a batch size of $32$ to train the gaze prediction network for a total of $50$ epochs.
For our motion forecasting network, the Adam optimiser with an initial learning rate of $0.01$ was used and the learning rate was decayed by $0.95$ every epoch.
A batch size of $32$ was employed to train the motion forecasting network for $100$ epochs.
Our method was implemented using the PyTorch framework.

\subsection{Motion Forecasting Results} \label{sec:motion_forecasting}




\def\arraystretch{1.2}

\begin{table}[!htbp]
        \caption{MPJPE errors (unit: millimeters) of different methods for motion forecasting on the MoGaze, ADT and GIMO datasets. Best results are in bold. 
        }\label{tab:global}
	\centering
	\resizebox{0.5\textwidth}{!}{
	\begin{tabular}{cccccccccccc}
		\toprule
		Dataset& Method & 200 ms & 400 ms & 600 ms & 800 ms & 1000 ms &Average\\ \hline
        \multirow{5}{*}{\textit{MoGaze}}
        &\textit{Res-RNN}~\cite{martinez2017human} &53.1 &91.3	&136.8 &187.5 &240.8 &124.3\\
        &\textit{siMLPe}~\cite{guo2023back} &40.6 &72.0 &108.8 &152.6 &201.0 &99.5\\
        &\textit{HisRep}~\cite{mao2021multi}  &31.4 &60.5 &95.4 &135.3 &177.9 &85.3\\
        &\textit{PGBIG}~\cite{ma2022progressively} &29.4 &57.7 &92.0 &130.7 &171.5 &82.0\\
        &Ours \textit{w/o gaze} &27.2 &55.3 &88.9 &126.9 &167.1 &79.0\\
        &Ours &\textbf{25.8}  &\textbf{53.3} &\textbf{85.8} &\textbf{122.0} &\textbf{160.0} &\textbf{75.9}\\
        \midrule
	\multirow{5}{*}{\textit{ADT}}
        &\textit{Res-RNN}~\cite{martinez2017human} &35.6 &55.7 &77.8 &100.0 &122.5 &70.1\\
        &\textit{siMLPe}~\cite{guo2023back}  &29.9 &48.3 &69.1 &93.8 &120.7 &63.8\\
        &\textit{HisRep}~\cite{mao2021multi}  &15.5  &30.5  &47.6  &66.8  &88.2 &42.3\\
        &\textit{PGBIG}~\cite{ma2022progressively} &14.5 &28.7  &45.4 &64.4  &85.8 &40.6\\
        &Ours w/o \textit{gaze} &12.0 &26.6 &44.0 &63.8	&85.3 &39.1\\
        &Ours &\textbf{11.7} &\textbf{25.8} &\textbf{42.8} &\textbf{62.1}  &\textbf{82.8} &\textbf{38.0}\\      
        \midrule
	\multirow{5}{*}{\textit{GIMO}}
        &\textit{Res-RNN}~\cite{martinez2017human} &82.6 &126.4 &170.2 &212.9 &255.4 &152.8\\
        &\textit{siMLPe}~\cite{guo2023back} &42.8 &78.3 &114.6 &150.7 &188.5 &100.3\\
        &\textit{HisRep}~\cite{mao2021multi}  &41.8  &78.1 &115.0  &152.7 &192.4 &100.2\\         
        &\textit{PGBIG}~\cite{ma2022progressively}  &38.0  &68.6  &101.9 &136.1  &172.2 &89.2\\
        &Ours w/o \textit{gaze} &33.7 &66.1 &99.7 &134.4 &170.4 &86.8\\
        &Ours  &\textbf{32.6}  &\textbf{64.1}	 &\textbf{97.0}  &\textbf{130.0} 	&\textbf{162.4} &\textbf{83.8}\\         
        \bottomrule        
	\end{tabular}} 
\end{table}

\begin{figure}
\centering
    \includegraphics[width=\columnwidth]{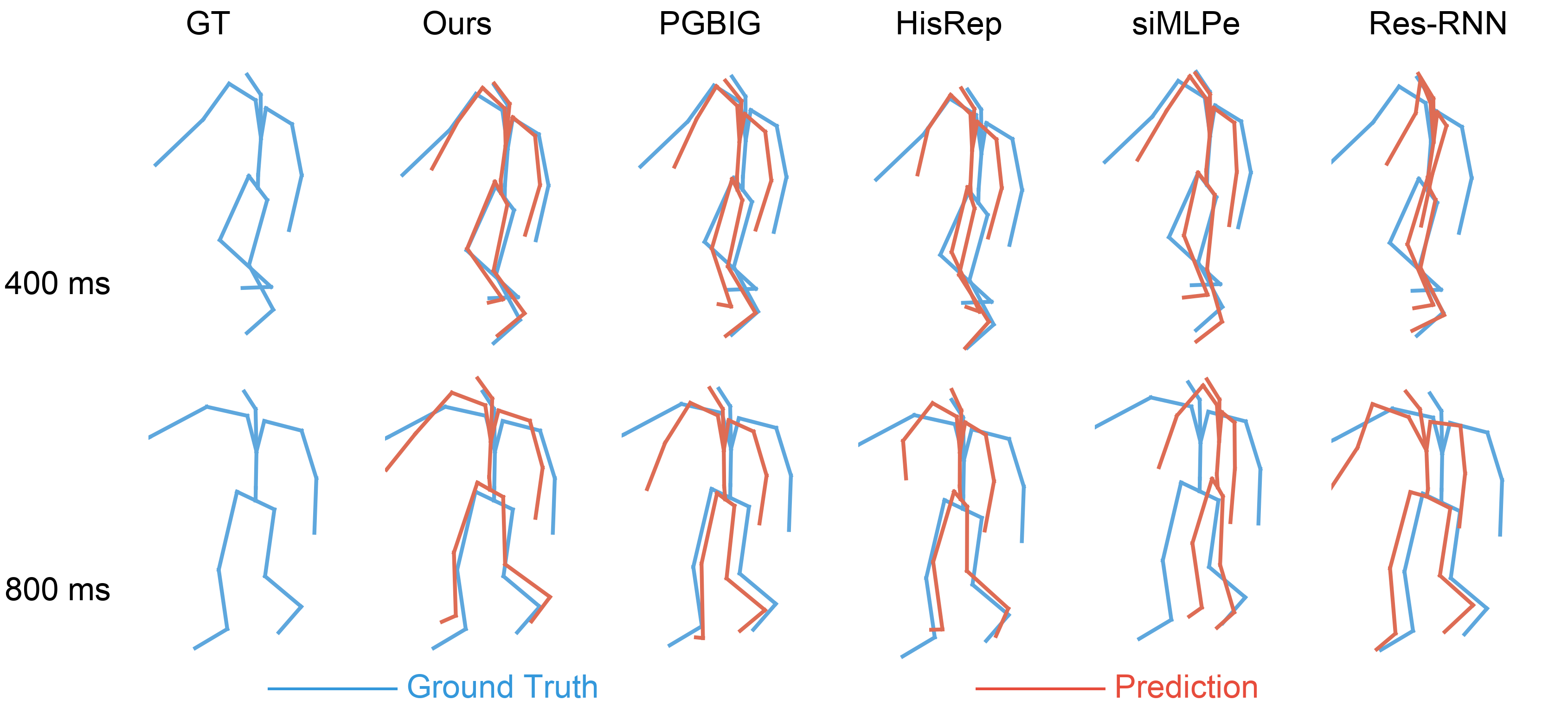}
    \caption{Visualisation of the predictions of different methods on MoGaze~\cite{kratzer2020mogaze}. Our method consistently outperforms other methods when predicting 400ms and 800ms into the future.}
    \label{fig:result}
\end{figure}

\textbf{Results on MoGaze}:
Table \ref{tab:global} summarises the performances of different methods on the MoGaze dataset.
The table shows the average MPJPE error (in millimeters) over all $30$ frames as well as the prediction errors for different future time horizons: $200$ ms, $400$ ms, \ldots, $1000$ ms.
As can be seen from the table, our method consistently outperforms the state-of-the-art methods at different future time intervals,
achieving an average improvement of $7.4\%$ ($75.9$ \textit{vs.} $82.0$) over the state of the art.
A paired Wilcoxon signed-rank test was used to compare the performances of our method with the state of the art and the results validated that the differences between our method and the state of the art are statistically significant ($p<0.01$).
Figure \ref{fig:result} shows an example of the predicted poses from different methods.
We can see that our method achieves significantly better performances than other methods.
See supplementary video for more prediction results.

\textbf{Results on ADT}:
Table \ref{tab:global} shows the MPJPE errors of different methods for different future time horizons as well as the average error.
We can see that our method significantly outperforms prior methods (paired Wilcoxon signed-rank test, $p<0.01$), achieving an overall average improvement of $6.4\%$ ($38.0$ \textit{vs.} $40.6$).

\textbf{Results on GIMO}:
We can see from \autoref{tab:global} that our method outperforms state-of-the-art methods with an overall average improvement of $6.1\%$ ($83.8$ \textit{vs.} $89.2$) and our improvements are statistically significant (paired Wilcoxon signed-rank test, $p<0.01$).

\def\arraystretch{1.2}

\subsection{Head Direction as a Proxy to Eye Gaze} \label{sec:head}

Our method requires eye gaze as input but gaze may not always be available, thus limiting the range of our method's possible applications.
To increase the usability of our method, we propose to use head direction (forward direction of head) as a proxy to eye gaze, inspired by the strong correlations between eye and head movements~\cite{sidenmark2019eye, hu2019sgaze}.
Specifically, we first predicted head direction from past head direction, then fused the predicted future head direction and past body poses, and finally applied our motion forecasting network to predict future motions.
Our method using head direction achieves an average improvement of $5.5\%$ ($77.5$ \textit{vs.} $82.0$), $5.2\%$ ($38.5$ \textit{vs.} $40.6$), and $3.8\%$ ($85.8$ \textit{vs.} $89.2$) over the state of the art on MoGaze, ADT, and GIMO, respectively.
These results demonstrate that our method can use head direction as a proxy to eye gaze and still obtain significantly better performances than the state of the art.

\subsection{Ablation Study}

\paragraph{Effectiveness of Eye Gaze for Motion Forecasting}
We tested different ways of using eye gaze information, i.e., 1) don't use eye gaze (\textit{w/o gaze}), 2) use past head direction in the gaze-pose fusion procedure (\textit{past head}), 3) use predicted future head direction (\textit{future head}), 4) use past eye gaze (\textit{past gaze}), and 5) use predicted future eye gaze (Ours).
Table \ref{tab:ablation_gaze} shows the motion forecasting results of different ways of using eye gaze on the MoGaze dataset.
As can be seen from the table, using eye gaze or head direction as a proxy to eye gaze achieves significantly better performances than not using eye gaze (paired Wilcoxon signed-rank test, $p<0.01$), validating that eye gaze information can help improve the performance of motion forecasting.
We also find that using predicted future eye gaze or head direction achieves better performances than directly using past eye gaze ($75.9$ \textit{vs.} $77.1$) or head direction ($77.5$ \textit{vs.} $77.8$), revealing the significant potential of applying gaze prediction methods to improve the performances of motion forecasting.

\def\arraystretch{1.2}
\begin{table}
        \caption{Ablation study on the MoGaze dataset.}\label{tab:ablation_gaze}
	\centering
	\resizebox{0.5\textwidth}{!}{
	\begin{tabular}{ccccccc}
		\toprule
		Method & 200 ms & 400 ms & 600 ms & 800 ms & 1000 ms &Average\\ \hline
         w/o \textit{spatial GCN} &30.9 &62.1 &96.3 &133.8 &173.1 &84.7\\
         w/o \textit{temporal GCN} &46.6 &74.0 &107.9 &147.0 &188.0 &99.3\\
         w/o \textit{copy} &26.8 &54.5 &87.1 &123.2 &161.0 &77.0\\
         w/o \textit{global residual} &32.9 &58.7 &90.7 &127.0 &165.0 &81.9\\
         w/o \textit{velocity loss} &26.3 &53.9 &86.6 &122.9 &161.1 &76.6\\
         \textit{w/o gaze} &27.2 &55.3 &88.9 &126.9 &167.1 &79.0\\
        \textit{past gaze} &26.3 &54.3 &87.2 &123.8 &162.0 &77.1\\

        \textit{past head} &26.3 &54.4	&87.9 &125.4 &164.1 &77.8\\
        \textit{future head} &26.1 &54.1 &87.5 &124.8 &163.7 &77.5\\\cline{1-7}
        Ours &\textbf{25.8} &\textbf{53.3} &\textbf{85.8} &\textbf{122.0} &\textbf{160.0} &\textbf{75.9}\\
        \bottomrule                    
	\end{tabular}}
\end{table}

\paragraph{Effectiveness of Our GCN Architecture}
From \autoref{tab:global} we can see that even without using eye gaze, our method still significantly outperforms the state of the art ($79.0$ \textit{vs.} $82.0$ on MoGaze, $39.1$ \textit{vs.} $40.6$ on ADT, $86.8$ \textit{vs.} $89.2$ on GIMO, paired Wilcoxon signed-rank test, $p<0.01$), validating the effectiveness of our GCN architecture.
Furthermore, we respectively removed \textit{spatial GCN}, \textit{temporal GCN}, \textit{copy} in start module, \textit{global residual}, and \textit{velocity loss}, and retrained the ablated methods.
We can see from \autoref{tab:ablation_gaze} that each component helps improve our method's motion forecasting performance.
We also find that \textit{temporal GCN} is most important to the success of our method, revealing the significance of temporal features for motion forecasting.

\paragraph{Training Parameters} 
We used different dropout rate to train our model and the performances of dropout rate $0.1$, $0.2$, $0.3$ (Ours), and $0.4$ on MoGaze are $77.7$, $76.9$, $75.9$, and $76.5$, respectively.
We also added a weighting constant for $\ell_v$ in \autoref{eq:loss} and the performances of weighting constant $0.25$, $0.5$, $1.0$ (Ours), and $1.5$ on MoGaze are $76.0$, $76.5$, $75.9$, and $76.3$, respectively.
These results validate that our training parameters are optimal in practice.

\subsection{User Study} \label{sec:user_study}

To further evaluate whether our method's improvements are significant in terms of qualitative evaluation, we conducted an online user study to compare our method with prior methods.

\subsubsection{Stimuli} 
We randomly selected $24$ motion forecasting samples from the MoGaze, ADT, and GIMO datasets ($8$ samples from each dataset) and used them as our stimuli.
Each sample consisted of $30$ frames of predictions (corresponding to future $1$ second) and was visualised as a short video.

\subsubsection{Participants}
We recruited 20 participants (12 males and 8 females, aged between
21 and 36 years) to take part in our user study through university mailing lists and social networks.
All of the participants reported normal or corrected-to-normal vision.
The user study was approved by our university's ethical review board.

\subsubsection{Procedure}
We conducted our user study using a Google form.
During the study, the ground truth future motions and the predictions of different methods were displayed to the participants in parallel using a layout that is similar to Figure \ref{fig:result}.
The names of different methods were hidden and the order of these methods were randomised.
The visualisation videos of the ground truth and different methods were set to loop automatically, allowing participants to observe them with no time limit.
During their observation, participants were required to rank different methods according to two criteria: \textit{precision} and \textit{realism}.
\begin{itemize}
\item \textit{Precision}: check different methods to see whether they \textit{align with the ground truth} and rank them based on your observation.
\item \textit{Realism}: check different methods to see whether they are \textit{physically plausible} and rank them based on your observation.
\end{itemize}
We collected the participants' responses
for further analysis.

\subsubsection{Statistical Analysis}
The means and standard deviations (SDs) of different methods' rankings are shown in Table \ref{tab:user_study}.
We can see that our method outperforms the state of the art in terms of both precision ($1.6$ \textit{vs.} $3.2$) and realism ($1.9$ \textit{vs.} $3.1$) and the results are statistically significant (paired Wilcoxon signed-rank test, $p<0.01$).
The above results demonstrate that our method achieves significantly better performances over prior methods in qualitative evaluation.

\begin{table}[t]
	\centering
        \caption{Statistical results of different methods' rankings in our user study.
        }\label{tab:user_study}
	\resizebox{0.48\textwidth}{!}{
	\begin{tabular}{ccccccc}
		\toprule
		 & & Ours & \textit{PGBIG}~\cite{ma2022progressively}& \textit{HisRep}~\cite{mao2021multi}& \textit{siMLPe}~\cite{guo2023back}& \textit{Res-RNN}~\cite{martinez2017human}\\ \hline
	\multirow{2}{*}{\textit{Precision}}
        & Mean & \textbf{1.6} & \underline{3.2} & \underline{3.2} & 3.3 & 3.7 \\
        & SD   & 0.9 & 1.2 & 1.2 & 1.3 & 1.3 \\ 
        \midrule
	\multirow{2}{*}{\textit{Realism}}
        & Mean & \textbf{1.9} & 3.3 & \underline{3.1} & 3.3 & 3.5 \\
        & SD   & 1.3 & 1.2 & 1.3 & 1.3 & 1.4 \\
        \bottomrule
	\end{tabular}}
\end{table}

\subsection{Discussion} \label{sec:discussion}

From the results in Table \ref{tab:global}, we find that the performances of all the methods deteriorate significantly with the increase of prediction time. This is a well-known problem in motion forecasting~\cite{ma2022progressively, guo2023back} that all current methods suffer from since these methods only use historical motion information.
Integrating more context information such as user's goal or task into motion forecasting has the potential to alleviate this problem.
In addition, we only explored the effectiveness of eye gaze and head direction on motion forecasting but ignored other important body signals such as hand gestures or gait.
Integrating such body signals into our pipeline to further improve the performance is an interesting avenue of future work.
Furthermore, we are also looking forward to incorporating stochasticity into the human motion forecasting model to further improve the performance.
\section{Conclusion}

In this work we proposed a novel method for human motion forecasting that first predicts future eye gaze from past gaze, fuses future eye gaze and past body poses into a gaze-pose graph, and finally uses a spatio-temporal residual GCN.
Through extensive experiments on three public benchmark datasets we showed that our method outperforms several state-of-the-art methods by a large margin.
We also validated that our predictions are more precise and more realistic than prior methods through an online user study.
We further showed that head direction can be a suitable proxy to eye gaze for use cases where eye gaze is not available, thereby further improving the applicability of our method.
As such, our work reveals the significant information content available in eye gaze for human motion forecasting and paves the way for future research on this promising research direction.

{
    \bibliographystyle{IEEEtran}
    \bibliography{references.bib}
}

\end{document}